# Advancing GDP Forecasting: The Potential of Machine Learning Techniques in Economic Predictions

Bogdan Oancea[1]*

[1]University of Bucharest, Faculty of Business and Administration, Department of Applied Economics and Quantitative Analysis, Bucharest, Romania

**Abstract:** The quest for accurate economic forecasting has traditionally been dominated by econometric models, which most of the times rely on the assumptions of linear relationships and stationarity in of the data. However, the complex and often nonlinear nature of global economies necessitates the exploration of alternative approaches. Machine learning methods offer promising advantages over traditional econometric techniques for Gross Domestic Product forecasting, given their ability to model complex, nonlinear interactions and patterns without the need for explicit specification of the underlying relationships. This paper investigates the efficacy of Recurrent Neural Networks, in forecasting GDP, specifically LSTM networks. These models are compared against a traditional econometric method, SARIMA. We employ the quarterly Romanian GDP dataset from 1995 to 2023 and build a LSTM network to forecast to next 4 values in the series. Our findings suggest that machine learning models, consistently outperform traditional econometric models in terms of predictive accuracy and flexibility.

## 1. Introduction

GDP forecasting is a critical process in economic planning and policymaking. Accurate GDP forecasts enable governments and financial institutions to make informed decisions regarding fiscal policies, budget allocations, and economic strategies. Traditional methods of GDP forecasting often rely on statistical models and historical data to identify trends and project future performance. However, these methods can sometimes fall short in capturing the complexities and nonlinearities of modern economies, especially during periods of rapid change or economic upheaval. In this context, economic forecasting in general and particularly GDP prediction has significantly benefited from the integration of machine learning techniques (Woloszko, 2017; Tamara et al., 2020).

Machine learning (ML) has emerged as a powerful tool in the realm of economic forecasting, offering significant advantages over traditional statistical methods. ML algorithms can process vast amounts of data and uncover intricate patterns and relationships that might not be apparent through conventional analysis. Techniques such as Random Forest, Support Vector Machines, or Neural Networks have been effectively employed to enhance the accuracy of GDP predictions and these methods can outperform traditional autoregressive models in forecasting GDP, providing more reliable and timely insights (Yoon, 2021; Chu & Qureshi, 2023). Thus, Tamara et al. (2020) examined the application of multiple machine learning algorithms, including Random Forest, LASSO and Ridge Regression, Elastic Net,

* Corresponding author's email: bogdan.oancea@faa.unibuc.ro

Neural Networks, and Support Vector Machines, for nowcasting Indonesia's GDP growth. They found that all machine learning models outperformed the autoregressive (AR) benchmark, with Random Forest showing the best individual performance. Combining forecasts using LASSO regression further improved accuracy, highlighting the effectiveness of ensemble approaches in economic forecasting. On the other hand, Flannery (2023) evaluates the accuracy of flash GDP estimates using deep learning approaches (LSTM models) compared to traditional time series econometric models (ARIMA, ARIMA with explanatory variables, and VAR), concluding that ARIMA models with explanatory variables offer the most accurate estimates and recommending the incorporation of additional explanatory variables for improving Ireland's flash GDP estimates. Other authors such as Vrbka (2016) applied neural networks to predict the GDP growth of Eurozone countries until 2025. The study utilized various neural network architectures and found that Radial Basis Function (RBF) networks provided the most accurate predictions. The neural network models showed strong correlation with actual GDP data, making them suitable for long-term economic forecasts. Kurihara and Fukushima (2019) compared traditional autoregressive (AR) models with machine learning models, specifically Long Short-Term Memory (LSTM) networks, for forecasting GDP and consumer prices in G7 countries. Their empirical results indicated that while traditional AR models slightly outperformed machine learning models in some cases, the latter showed substantial potential, particularly for data with strong trends. In another work, Paruchuri (2021) examined the application of machine learning in forecasting the Italian economy. The study employed various machine learning techniques, including nonlinear autoregression (NAR), support vector regression (SVR), and boosted trees (BT). The results underscored that machine learning models, especially NARX and SVR, provided quick and reliable predictions, essential for timely economic policy adjustment.

In this paper we propose a special type of recurrent neural networks, namely the Long Short-Term Memory (LSTM) networks to forecast the GDP in a univariate setting, and compare its performances with an econometric model, SARIMA, showing that the ML method outperform the econometric approach.

The rest of the paper is organized as follows. In the next section, we present the data series used in our experiment, followed by the proposed neural network model for GDP forecasting. A special section is dedicated to the presentation of the results, where we compare the performance of the neural network with the SARIMA method. We end our paper with a conclusion section.

## 2. Data

We used the quarterly GDP data from Q1 1995 to Q4 2023, CAEN Rev.2 gross series, average prices of 2020, values given in national currency, for Romania. The GDP data in the dataset provide a comprehensive view of economic performance over nearly three decades. The descriptive statistics for the period under analysis are reported in Table 1 while figure 1 shows the evolution of the GDP. The data was retrieved from the National Statistics Institute of Romania's database (http://statistici.insse.ro:8077/tempo-online/#/pages/tables/insse-table)

Table 1: Descriptive statistics

| Statistics | GDP |
|---|---|
| Mean | 199681.0552 |
| Standard Error | 6114.13206 |
| Median | 192018.65 |
| Standard Deviation | 65851.21759 |
| Kurtosis | -0.483138316 |
| Skewness | 0.389917727 |
| Range | 274466.4 |
| Minimum | 84817.9 |
| Maximum | 359284.3 |

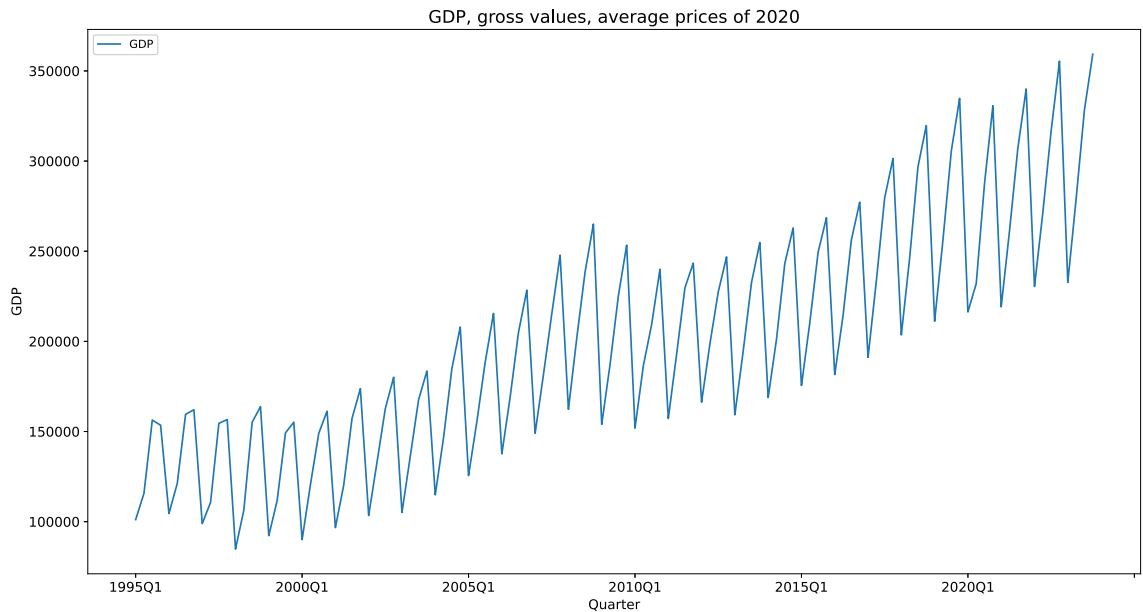

Figure 1: Quarterly GDP of Romania, gross series

As can be easily seen, the data series has a pronounced seasonality and an ascending trend, making it non-stationary. The ADF statistics is 0.31, with p-value=0.97 confirming the non-stationarity hypothesis.

Machine learning (ML) methods for time series forecasting are inherently flexible and robust, making them well-suited to handle non-stationary data. Unlike traditional statistical techniques, such as ARIMA, which require the data to be stationary to make accurate predictions, many ML algorithms do not impose any requirement on the data sets. ML models, including neural networks, decision trees, or ensemble methods, can learn and adapt to complex patterns in the data, including trends and seasonality, without the need for explicit differencing or detrending. These models can automatically capture and adjust to the underlying data dynamics, even when the statistical properties of the series change over time.

## 3. Methods

In this paper we propose a special type of recurrent neural networks, namely the Long Short-Term Memory (LSTM) networks (Hochreiter & Schmidhuber, 1997) to be used for GDP forecasting. LSTM networks have gained prominence for their effectiveness in time series forecasting. LSTMs are designed to capture temporal dependencies in data, making them particularly suitable for sequential data where the order of observations is crucial as it is the case with time series. Traditional RNNs suffer from the vanishing gradient problem, which hampers their ability to learn long-term dependencies. LSTMs address this issue with their unique architecture that includes cell states and gating mechanisms, allowing them to maintain and update information over long periods. This makes LSTMs especially adept at handling time series data, where understanding the relationship between past and future observations is essential for accurate forecasting (Graves et al., 2009).

To show the advantages of using ML methods for time series forecasting we compared the results obtained with the LSTM networks with the ones obtained with a classic econometric model, SARIMA (Korstanje, 2021).

In order to use the LSTM for GDP forecasting we divided our GDP series in two subsets for training and testing purposes. We kept the last 4 values in the series (Q1:2023 – Q4:2023) as testing values while the rest of the data were used for training the models. No other preprocessing operations were performed on the data series.

For the implementation of the LSTM networks, we used Keras (Chollet et al., 2015), and TensorFlow (Abadi et al., 2016), libraries. These libraries offer multiple advantages such as scalability, flexibility and easy-of-use. The SARIMA model was also implemented using Python together with the statsmodel library (Seabold & Perktold, 2010). The neural network model used in our study has 2 LSTM layers and one final dense layer. We used the ReLU activation function, the *adam* optimizer and Mean Squared Error as loss function. To avoid overfitting, we used regularization techniques such as dropout of the recurrent neurons in the second LSTM layer and an L2 kernel regularization in the final dense layer.

Hyperparameter tuning is very important for optimizing machine learning models used in economic time series forecasting. Both classical econometric methods and advanced deep learning architectures depend heavily on hyperparameters to perform optimally. Selecting the right hyperparameters greatly impacts the model's predictive accuracy, robustness, and ability to capture the complex patterns inherent in economic data. To select the best values for the hyperparameters, we employed a grid search for both models.

Table 2: The search space for the hyperparameters

| SARIMA | | LSTM network | |
| --- | --- | --- | --- |
| Parameter | Values | Parameter | Values |
| p, d, q | 0,1,2 | Training epochs | 250, 500,1000 |
| P, D, Q | 0, 1 | Recurrent dropout | 0.0, 0.1, 0.2, 0.3 |
| s | 4 (quarterly data) | Number of neurons in the LSTM layers | 250, 500,1000 |
| | | Batch size | 1,4, 8 |
| | | L2 for the kernel regularizer | 0.01, 0.02, 0.03 |

## 4. Results

For the grid search procedure applied to the LSTM network we used a 3-fold cross validation while for SARIMA we used the AIC score to select the best values for the parameters. These values are presented in table 3.

Table 3: The best values of the hyperparameters for SARIMA and LSTM models

| SARIMA | | LSTM network | |
|---|---|---|---|
| Parameter | Best Values | Parameter | Best Values |
| p, d, q | 2 | Training epochs | 1000 |
| P, D, Q | 1 | Recurrent dropout | 0.1 |
| s | 4 | Number of neurons in the LSTM layers | 250 |
| | | Batch size | 1 |
| | | L2 for the kernel regularizer | 0.01 |

After building and training the models with the best hyperparameter values we used them to predict the values from the test data set and computed the Mean Squared Error (MSE), Mean Absolute Error (MAE) and Mean Absolute Percentage Error (MAPE) performance metrics.

In figure 2 we present the actual versus the predicted values for the training and testing data sets for the LSTM network while in figure 3 we present the performance metrics monitored during the training process.

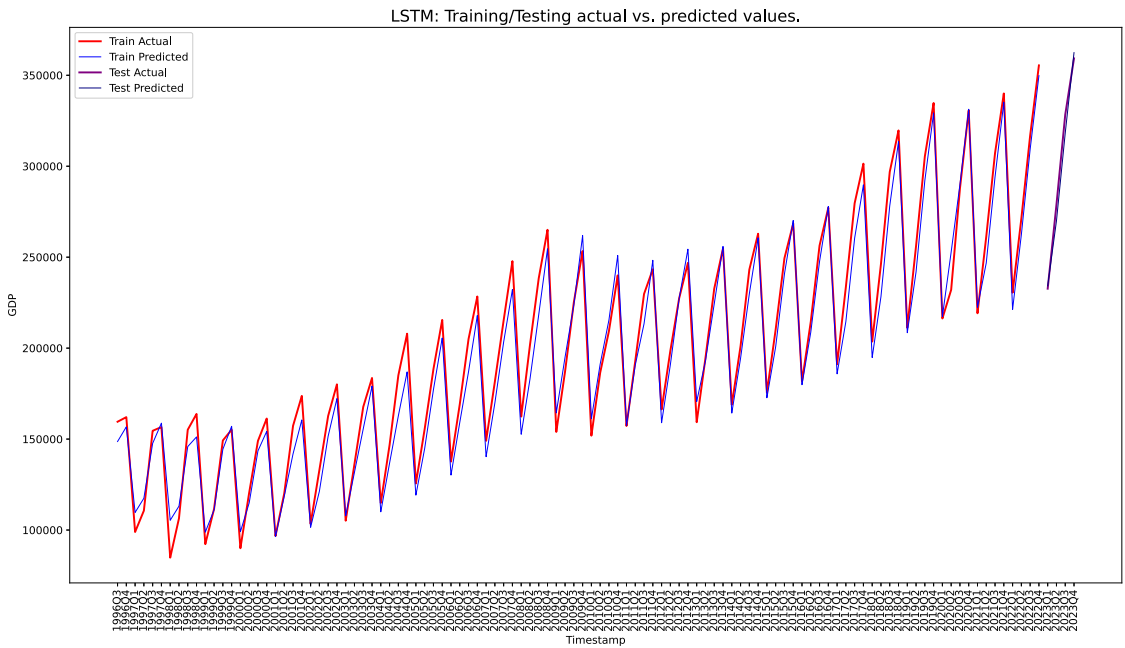

Figure 2: Training/testing actual versus predicted values

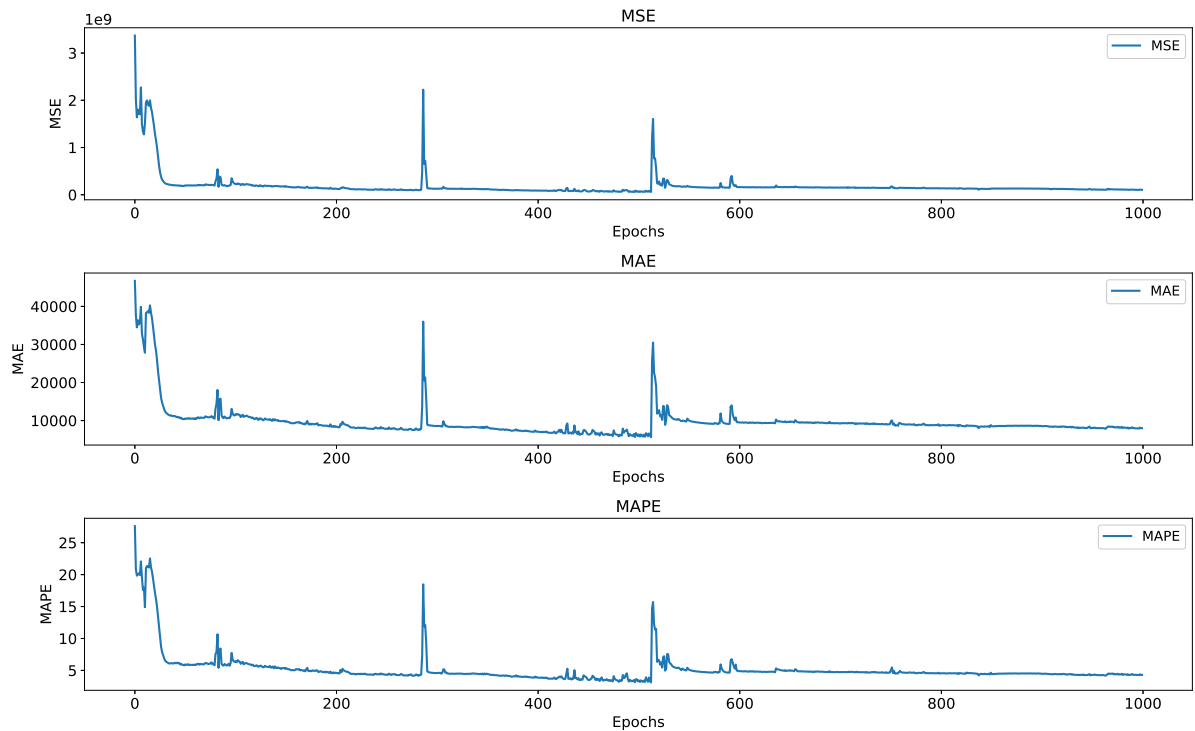

Figure 3: MSE, MAE and MAPE for the LSTM network during training

The performance metrics for the test data set, for both SARIMA and LSTM are presented in table 4.

Table 4: Performance metrics for the SARIMA and LSTM models

| Performance metrics | LSTM | SARIMA |
|---|---|---|
| MAPE | 1.96% | 2.56% |
| MAE | 5955 | 7144 |
| MSE | 48670983 | 63999869 |

These results show that LSTM network outperforms the SARIMA model for all performance metrics and align with numerous studies that demonstrate the superiority of machine learning methods over traditional econometric techniques for GDP forecasting. By leveraging advanced algorithms and the ability to model complex, nonlinear relationships in data, ML methods consistently provide more accurate and robust forecasts. This is particularly evident in our findings, where LSTM network outperformed traditional econometric models, such as SARIMA, in terms of predictive accuracy and the ability to capture intricate patterns in the economic data. These results corroborate existing research, underscoring the potential of ML approaches to enhance the precision and reliability of GDP forecasts, thereby offering significant advantages for economic analysis and policymaking.

## 5. Conclusion

Machine learning has proven to be a powerful tool for economic forecasting, offering improved accuracy and the ability to handle complex, nonlinear data. Studies indicate that techniques like Random Forest, Adaptive Trees, and neural networks are particularly effective for GDP prediction. In this study we compared the performances of a LSTM network with a SARIMA model showing that the

LSTM has a better performance for forecasting the GDP, all metrics computed for our experiment showing better values for LSTM compared with SARIMA.